%% file: main.tex
 \documentclass[tablecaption=bottom,wcp]{jmlr} 

\theorembodyfont{\upshape}
\theoremheaderfont{\scshape}
\theorempostheader{:}
\theoremsep{\newline}

\input{includes}

\jmlrproceedings{AABI 2025}{Workshop at the 7th Symposium on Advances in Approximate Bayesian Inference (non-archival), 2025}

\title[Towards One Model for Classical DR]{Towards One Model for Classical Dimensionality Reduction: A Probabilistic Perspective on UMAP and t-SNE}

\author{\Name{Aditya Ravuri} \nametag{\thanks{aditya.ravuri@cl.cam.ac.uk}} \and
 \Name{Neil D. Lawrence}\\
 \addr University of Cambridge}

\begin{document}

\maketitle

\vspace{-1cm}
\input{body}


\bibliography{main}

\appendix
\input{appendix}

\end{document}

%% file: includes.tex
\usepackage{amssymb}
\usepackage[american]{babel}
\usepackage{microtype}
\usepackage{graphicx}
\usepackage{booktabs}
\usepackage{amsmath}
\usepackage{bm}
\usepackage{amssymb}
\usepackage{mathtools}
\usepackage{multicol}
\usepackage{lastpage}
\usepackage{tikz}
\usepackage{changepage}
\usepackage{soul}
\usepackage{cleveref}
\usepackage{float}
\usepackage{caption}

\usetikzlibrary{bayesnet}

\newcommand{\Y}{\mathbf{Y}}

\newcommand{\M}{\mathbf{M}}
\newcommand{\X}{\mathbf{X}}
\newcommand{\A}{\mathbf{A}}
\renewcommand{\L}{\mathbf{L}}
\newcommand{\I}{\mathbf{I}}
\renewcommand{\S}{\mathbf{S}}
\renewcommand{\H}{\mathbf{H}}

\newcommand{\C}{\mathbf{C}}

\renewcommand{\C}{\mathbf{C}}

\newcommand{\q}{q} 
\newcommand{\kt}{\Tilde{k}}
\renewcommand{\P}{\mathbf{P}}

%% file: body.tex
\begin{abstract}
This paper shows that dimensionality reduction methods such as UMAP and t-SNE, can be approximately recast as MAP inference methods corresponding to a model introduced in \cite{probdr}, that describes the graph Laplacian (an estimate of the data precision matrix) using a Wishart distribution, with a mean given by a non-linear covariance function evaluated on the latents. This interpretation offers deeper theoretical and semantic insights into such algorithms, and forging a connection to Gaussian process latent variable models by showing that well-known kernels can be used to describe covariances implied by graph Laplacians. We also introduce tools with which similar dimensionality reduction methods can be studied.
\end{abstract}

\section{Introduction}
\label{sec:intro}

In domains with complex, high-dimensional data, such as single-cell biology, dimensionality reduction (DR) algorithms are essential tools for uncovering the underlying structure of data. These algorithms, which include very widely-used techniques like t-SNE \citep{tsne} and UMAP \citep{umap}, are especially valuable for visualizing data manifolds, enabling downstream processing, and the discovery of insightful clusters and trajectories. A deeper understanding of these algorithms and their theoretical underpinnings is crucial for advancing their applicability (particularly when prior information is available) and improving their interpretability. Our work builds on and aims to unify the Probabilistic Dimensionality Reduction (ProbDR) framework \cite{probdr}, which interprets classical DR methods through a probabilistic lens to enable the communication of assumptions, integration of prior knowledge, and model extension for new applications.

\cite{probdr} introduced ProbDR as a framework with two main interpretations:
\begin{enumerate}
    \item UMAP and t-SNE having a variational interpretations, describing inference over a nearest neighbour adjacency matrix, and,
    \item other classical eigendecomposition-based DR algorithms as inference algorithms of a Wishart model that describes a covariance/precision matrix with a linear kernel evaluated on the latents.
\end{enumerate}

\textbf{Our contribution:} In this work, we simplify the framework, moving away from the variational interpretations and propose that \textbf{all algorithms} with ProbDR interpretations (and hence most classical DR methods) in the large-$n$ limit can be written as MAP inference algorithms given the model \footnote{Note that we assume an improper prior on the latents, $p(\X) \propto 1$, throughout this work.},
\begin{equation}
    \label{eqn:probdr}
    \mathbf{S} | \mathbf{X} \sim \mathcal W^{(-1)} ( \mathbf{XX}^T + \alpha \H K_{t}(\mathbf{X}) \H + \gamma \mathbf{I}, \nu),
\end{equation}
where $\S \in S^+_n$ is an estimate of a covariance matrix generated using the high-dimensional data $\Y \in \mathbb{R}^{n, d}$, $\X \in \mathbb R^{n, \q}$ corresponds to the set of low ($\q$-)dimensional latent variables, and $K$ is a covariance function used to construct a positive-definite matrix constructed using latent variables.
This enables a direct comparison between algorithms such as Laplacian Eigenmaps, t-SNE and UMAP.\footnote{Moreover, certain constructions used in this paper, such as double centered distance matrices and exponentiated kernel matrices may be useful for building covariances in practice.}

The rest of the paper will take the following structure. In section 2 we outline relevant background work that we build on in developing the unified framework which we derive in section 3. In section 4, we discuss the significance of this framework in the ability it affords in comparing UMAP and t-SNE based on differences in kernel type. We also present empirical results and connections to Gaussian process latent variable models \cite{gplvm}. Section 5 concludes the paper.

\section{Background}
\label{sec:background}

We will now discuss related background work which we build on in developing the unified framework in the next section. Our main objective is to simplify the ProbDR framework by moving away from its variational interpretations\footnote{\cite{assel} also offer a similar \textit{variational} view on UMAP and t-SNE.} and to instead consider algorithms such as t-SNE and UMAP as MAP algorithms. If a Wishart model was found, inference wherein approximated t-SNE and UMAP, assumptions within that model could be compared to the model that leads to traditional eigendecomposition-based methods such as Laplacian Eigenmaps \citep{lap-eigenmaps}.

For this work, we need to recap only the eigendecomposition view of \cite{probdr}, and we do so with \textbf{Laplacian Eigenmaps} as an example: the algorithm involves the calculation of a nearest neighbour graph using the high-dimensional datapoints $\Y$, which can be represented using a graph Laplacian $\hat{\L} = \mathbf{D} - \A$ (where $\A$ is the corresponding adjacency matrix, and $\mathbf{D}$ is the diagonal degree matrix, $\mathbf{D}_{ii}$ = $\sum_k \A_{ik}$) and then obtaining the embedding $\X$ as the eigenvectors of $\hat{\L}$ corresponding to the lowest eigenvalues.

ProbDR showed that this corresponds to inference for $\X$ by maximising $\log \mathcal{W}(\nu\hat{\L} | (\X\X^T + \beta \I)^{-1}, \nu)$, i.e. maximising the likelihood of,
\begin{equation}
\label{eqn:le}
\nu*\hat{\L} | \X \sim \mathcal{W}((\X\X^T + \beta \I)^{-1}, \nu),
\end{equation}
where $\hat{\L}$ is interpreted to be an estimate of a precision matrix (see \cite{probdr} for more detail).\footnote{Although the proof method there uses results used by \cite{mca} (for probabilistic minor components analysis), a similar result can be seen easily. Consider the matrix $\hat{\S} = (\hat{\L} + \epsilon \I)^{-1}$. Maximising the likelihood of $d*\hat{\S} | \X \sim \mathcal W(\X\X^T + \beta\I, d)$ recovers the same solution as the one above. This can be seen as the major eigenvectors of $\hat{\S}$ are the minor eigenvectors of $\hat{\L}$. This model is simply dual probabilistic PCA \citep{gplvm}, but with a non-standard covariance estimator, instead of the traditional $\Y\Y^T/d$, written in terms of the covariance, as $\Y \sim \mathcal{MN}(\boldsymbol{0}, \C, \I) \Rightarrow \Y\Y^T \sim \mathcal{W}(\C, d)$.} The model is intuitive as the implied covariance ($\hat{\L}^+$) is modelled by a linear covariance function acting on the latents $\X$, which is familiar in models such as (dual probabilistic) PCA and GPLVMs \citep{gplvm}.


UMAP and t-SNE, in \cite{probdr}, were interpreted in a variational way, i.e. as KL-minimising algorithms acting on the binary adjacency matrix $\A$. Under certain circumstances\footnote{if the variational probabilities are zero or one, signifying if two points are nearest neighbours; this simplification is reasonable due to the findings of \cite{umap_loss}, where it was found that the relatively complex calculation of the variational probabilities in t-SNE and UMAP can be replaced with simply the adjacency matrices without loss of performance. Our initial experiments closely aligned with these findings.}, the interpretation becomes equivalent to MAP estimation (due to \cite{probdr}, Appendix B.7, Lemma 13, also presented in \cite{nc_sne}).

Critically, however, \cite{nc_sne} note that the optimisation process is equally as important. As part of an extensive study on the nature of the t-SNE and UMAP loss functions, \cite{nc_sne} then show how the stochastic optimisation of t-SNE and UMAP can be interpreted to be contrastive estimation with the energy function (negative loss),
\begin{align}
\label{eqn:cne-bound}
    \mathcal{E}(\X) \propto &\sum_{ij} \A_{ij} \log\left(\frac{1}{d_{ij}(\X)^2 + 1}\right) + \dfrac{4n_{\text{neg}} n_{\text{neigh}}}{3n}\sum_{ij} (1 - \A_{ij}) \log\left(1 - \frac{1}{d_{ij}(\X)^2 + 1}\right),
\end{align}
where $\A_{ij}$ represents whether data points $\Y_i$ and $\Y_j$ are neighbours, and $d_{ij}^2(\X) = \| \X_{i:} - \X_{j:} \|^2$. We will refer to this as the \textbf{CNE objective}. Note that we've made some substitutions in the original formation that appears in \cite{nc_sne}, for example, we set their parameter $\Tilde{q}_{ij} = 1/d_{ij}^2$, which corresponds to the UMAP setting (and a full derivation is given in \cref{app:derivation-cne-bound}). The hyperparameter $n_{\text{neg}}$ sets the number of contrastive negatives (set to be five) that affects the strength of repulsion, and $n_{\text{neigh}}$ corresponds to the number of neighbours set for a point (fifteen in this work). In this work, we will work with this loss function and aim to interpret it as a likelihood, but over the latents $\X$.\footnote{We were particularly inspired to look toward contrastive methods by \cite{ssl_as_vi}, who showed that contrastive learning methods could be seen as variational algorithms (hence suggesting a link between t-SNE/UMAP and contrastive learning) and by \cite{nce}, which shows that contrastive losses are estimators of negative log-likelihoods. \cite{nc_sne} offer us a constrastive objective for t-SNE and UMAP, greatly simplify the optimisation process and offer a clear objective to work with, i.e. \cref{eqn:cne-bound}.}

\section{Towards a distribution on the graph Laplacian}

In this section, we derive the Wishart distribution inference in which leads to UMAP and t-SNE-like algorithms.
Eq.~\ref{eqn:cne-bound} is not a likelihood due to the multiplicative constant weighting the contribution of points that are not adjacent. The second term, $\mathcal T_b$ can be expressed as follows, where $\Tilde{\epsilon} = 4n_{\text{neg}} n_{\text{neigh}}/3n$,
\begin{align*}
    \mathcal T_b = &\sum_{ij} (1 - \A_{ij}) \log\left( \left[ 1 - \frac{1}{d_{ij}(\X)^2 + 1} \right]^{\Tilde{\epsilon}} \right).
\end{align*}

The implied probability of adjacency $\Tilde{\mathbf{p}}_{ij}$, assuming that this is the second term of a Bernoulli likelihood is,
\begin{align*}
    \Tilde{\mathbf{p}}_{ij} &= 1 - \left[ 1 - \frac{1}{d_{ij}(\X)^2 + 1} \right]^{\Tilde{\epsilon}} = 1 - \exp\left[ \Tilde{\epsilon} \log \left( 1 - \frac{1}{d_{ij}(\X)^2 + 1} \right) \right] \\
    &= 1 - \exp\left[ -\Tilde{\epsilon} \log \left(1 + \frac{1}{d_{ij}(\X)^2} \right) \right] \approx 1 - 1 + \Tilde{\epsilon} \log \left(1 + \frac{1}{d_{ij}(\X)^2} \right) && \text{large n}.
\end{align*}

Now, we observe that $\log (1 + 1/x) >= 1/(1+x)$, so,
\begin{align*}
    \mathcal{T}_b &\approx \sum_{ij} (1 - \A_{ij}) \log \left(1 - \Tilde{\mathbf{p}}_{ij}  \right) \leq \sum_{ij} (1 - \A_{ij}) \log \left(1 - \Tilde{\epsilon} \dfrac{1}{1 + d_{ij}(\X)^2} \right).
\end{align*}

Note that the multiplication of $\Tilde{\epsilon}$ within the log in the first term adds just a constant to the first term of \cref{eqn:cne-bound}). Therefore,
\begin{align*}
    \mathcal{E}(\X) \leq &\sum_{ij} \A_{ij} \log\left(\Tilde{\epsilon} \frac{1}{1 + d_{ij}(\X)^2}\right) + \sum_{ij} (1 - \A_{ij}) \log \left(1 - \Tilde{\epsilon} \dfrac{1}{1 + d_{ij}(\X)^2} \right) + c.
\end{align*}

We conclude that the CNE objective lower bounds the Bernoulli likelihood implied by the model,
\begin{equation}
    \label{eqn:bern-interp}
    \A_{ij} | \X \sim \text{Bernoulli} \left( \Tilde{\epsilon} \dfrac{1}{1 + d_{ij}(\X)^2} \right).
\end{equation}

Although this is a valid probabilistic interpretation, we will go slightly further and outline an argument that makes this interpretation comparable to the Wishart interpretations in ProbDR.\footnote{This is reasonable as exponential families share similar likelihood forms, and a Wishart interpretation, despite being over discrete matrices, may correspond to a similar statistical learning scenario as performing classification using linear regression (i.e. using the $L^2$ norm serparator).} As Wisharts have supports over positive definite matrices, we will try to reconsider this model with the graph Laplacian $\L$ as the observed statistic. The likelihood for $\X$ implied by \cref{eqn:bern-interp} is,
\begin{align*}
    \log p(\X | \A) = &\sum_{ij} \A_{ij} \log\left(\Tilde{\epsilon} \frac{1}{1 + d_{ij}(\X)^2}\right) + \sum_{ij} (1 - \A_{ij}) \log \left(1 - \Tilde{\epsilon} \dfrac{1}{1 + d_{ij}(\X)^2} \right) \\
    \approx &\sum_{ij} \A_{ij} \log\left(\Tilde{\epsilon} \frac{1}{1 + d_{ij}(\X)^2}\right) + \sum_{ij} \log \left(1 - \Tilde{\epsilon} \dfrac{1}{1 + d_{ij}(\X)^2} \right) && \text{large n} \\
    \approx & \; \text{tr}(\A \P') - \sum_{ij} \Tilde{\epsilon} \dfrac{1}{1 + d_{ij}(\X)^2} && \text{small } \Tilde{\epsilon}
\end{align*}
\begin{align*}
    \Rightarrow \log p(\X | \L) = & -\text{tr}(\L \H \P' \H) - \sum_{ij} \Tilde{\epsilon} \dfrac{1}{1 + d_{ij}(\X)^2}, && \text{centd. } \L \text{ and tr}(\mathbf{D}\P') = 0
\end{align*}
where,
$\P^u_{ij} = 1/(1 + d_{ij}^2)$, $\P_{ij} = \Tilde{\epsilon}\P^u_{ij}, \; \P'_{ij} = \log \P_{ij}$\footnote{$\P'$ is conditionally positive definite (CPD) as $\P_{ij}$ defines the student-t kernel, which enforces PDness and an elementwise log of a PD matrix with all positive elements will at least be CPD (\cite{psd-mats}). Double centering such matrices makes them PSD.} and $\H = \I - \mathbf{1}\mathbf{1}^T/n$ is a centering matrix. Note that $\text{tr}(\H \M \H) = n - \sum_{ij} \M_{ij}/n$. Therefore\footnote{Note that our derivation also produces a similar objective to the DK-LLE objective of \cite{dk-lle} (Lemma 4), which was found by gradient-based arguments, adding credibility to our arguments.},
\begin{align*}
    \log p(\X | \L) &= -\text{tr}(\L \H \P' \H) + n \text{tr}(\H \P \H) + k \approx -\text{tr}(\L \H \P' \H) + n \text{log}| \I + \H \P \H | +k. && \text{small } \Tilde{\epsilon}
\end{align*}

We approximate $\log \P_{ij}$ within the vicinity of $\Tilde{\epsilon}$\footnote{This neighbourhood was chosen as $\Tilde{\epsilon}$ is the maximal value of $\P_{ij}$.} by matching the gradient and function value using an ansatz,
\begin{align*}
    \log \P_{ij} &\approx \dfrac{\P_{ij}}{2 \Tilde{\epsilon}} - \dfrac{\Tilde{\epsilon}}{2\P_{ij}} + \log{\Tilde{\epsilon}}. \\
    \Rightarrow \log p(\X | \L) &\approx -\text{tr}(\L \H (0.5\P^u - 0.5[1/\P^u]_{ij}) \H) + n \text{log}| \I + \H \P \H | +c \\
    &= -\text{tr}(\L (0.5\Tilde{\epsilon}^{-1}\I + 0.5 \H \P^u \H + \X \X^T)) + n \text{log}| 0.5\Tilde{\epsilon}^{-1}\I + 0.5\H \P^u \H | + k \\
    &\leq \log \mathcal W(\L|(0.5\Tilde{\epsilon}^{-1}\I + 0.5\H \P^u \H + \X \X^T)^{-1}, n).
\end{align*}

Therefore, the model below approximates the model in \cref{eqn:bern-interp}.
\begin{align}
    \label{eqn:wish-interp}
    \L | \X \sim \mathcal{W} \left((0.5\Tilde{\epsilon}^{-1}\I + 0.5 \H \P^u \H + \X \X^T)^{-1}, n \right)
\end{align}

\section{Results and Discussion}

In this paper, we've presented an interpretation for UMAP and t-SNE-like algorithms and proposed a comparable distributional assumption to ProbDR's. Our model for UMAP and t-SNE-like algorithms uses a covariance based on a double-centred non-linear kernel\footnote{Other empirical arguments can also be made to show that, assuming a model such as: $\X \rightarrow \Y \rightarrow \A$, where $\Y | \X \sim \mathcal{N}$, the adjacency probabilities of \cref{eqn:bern-interp} cannot be found using simply a linear kernel, and \textbf{can} be found using a non-linear kernel such as a Student-t kernel, as we've used here. For this, one can write out the approximate distribution on distances (assuming a Gaussian process prior on $\Y$, $\mathbb E(d_{ij}^2(\Y)) = d * (k_{ii} + k_{jj} - 2k_{ij}) \text{ and } \text{Cov}(d_{ij}^2, d_{mn}^2) = 2d * (k_{im} + k_{jn} - k_{in} - k_{jm})^2$) and study the probability with which such distances attain extreme values, as a function of latent distance.} however, as opposed to the linear kernel used as part of ProbDR's Wishart-based algorithms. Our work provides shows a direct connection to the model behind Laplacian Eigenmaps; the model of \cref{eqn:wish-interp} is a non-linear extension (albeit with an interesting choice of kernel scales) of ProbDR's Laplacian Eigenmaps with a non-linear kernel in \cref{eqn:le}.

\begin{figure*}[htp]
\centering
\begin{tabular}{ccc}
\begin{minipage}{0.3\textwidth}
    \centering
    \includegraphics[width=\textwidth]{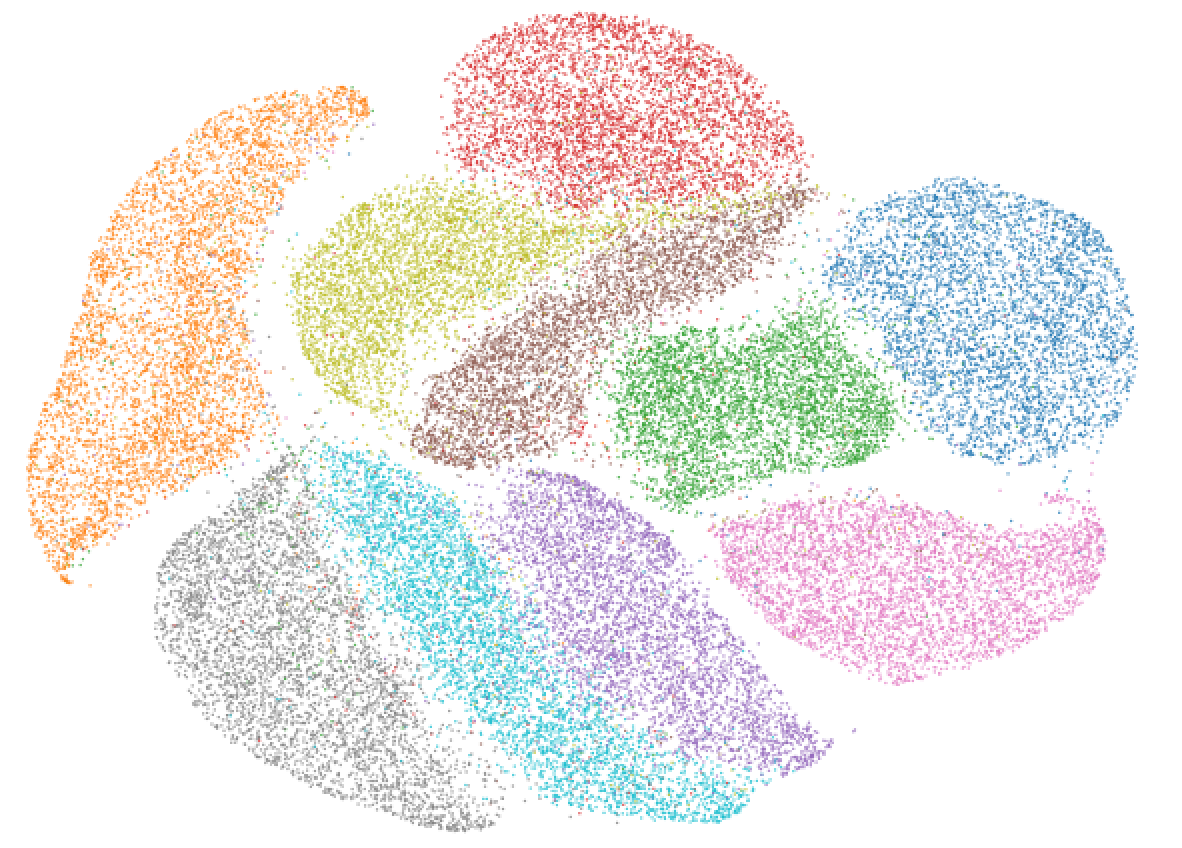}
\end{minipage} &
\begin{minipage}{0.3\textwidth}
    \centering
    \includegraphics[width=\textwidth]{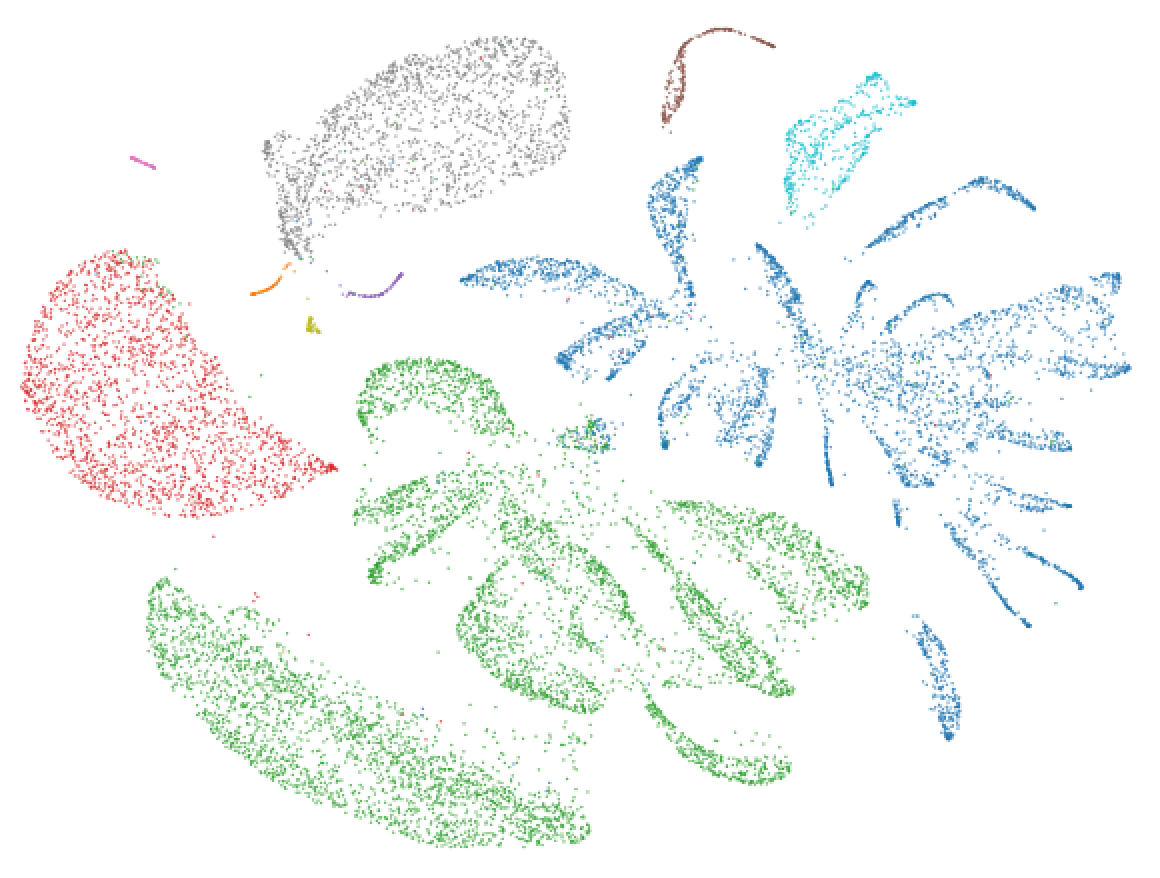}
\end{minipage} &
\begin{minipage}{0.3\textwidth}
    \centering
    \includegraphics[width=\textwidth]{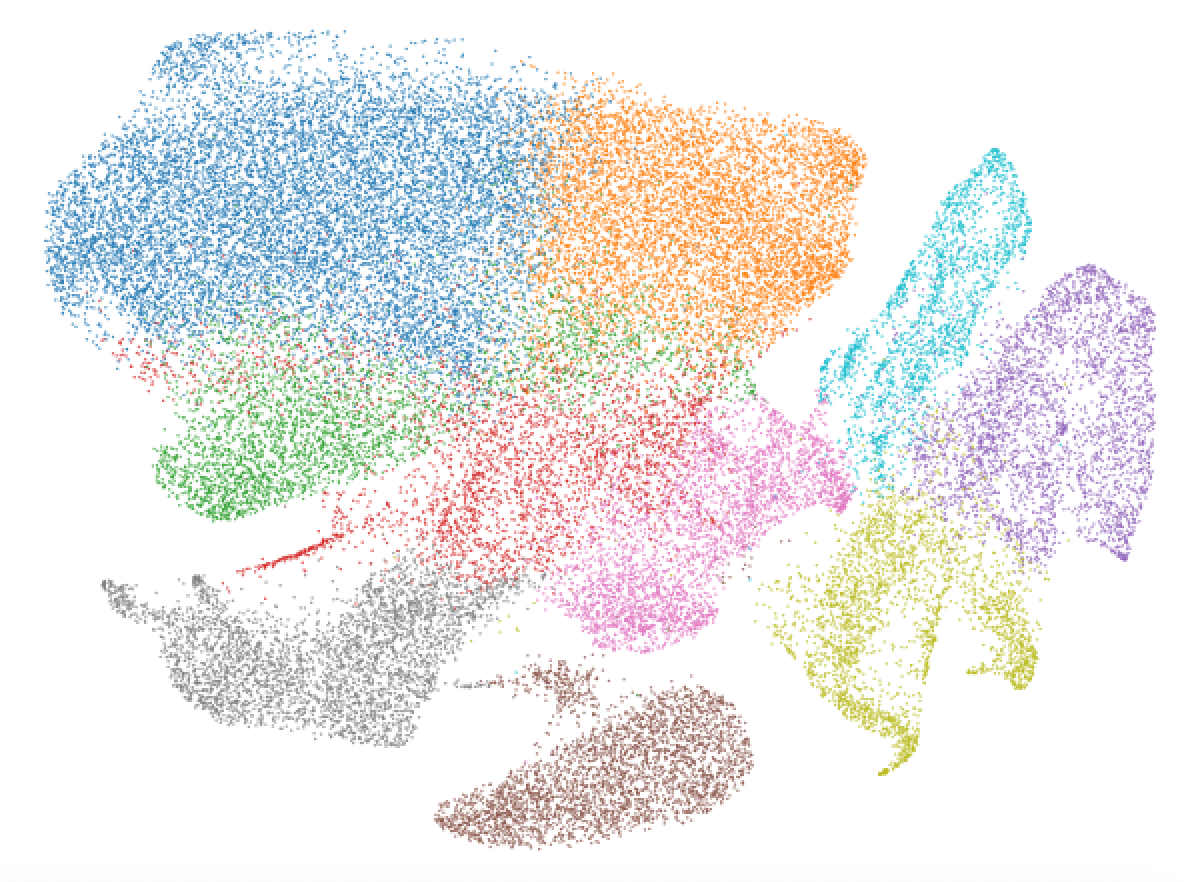}
\end{minipage} \\
\begin{minipage}{0.3\textwidth}
    \centering
    \includegraphics[width=\textwidth]{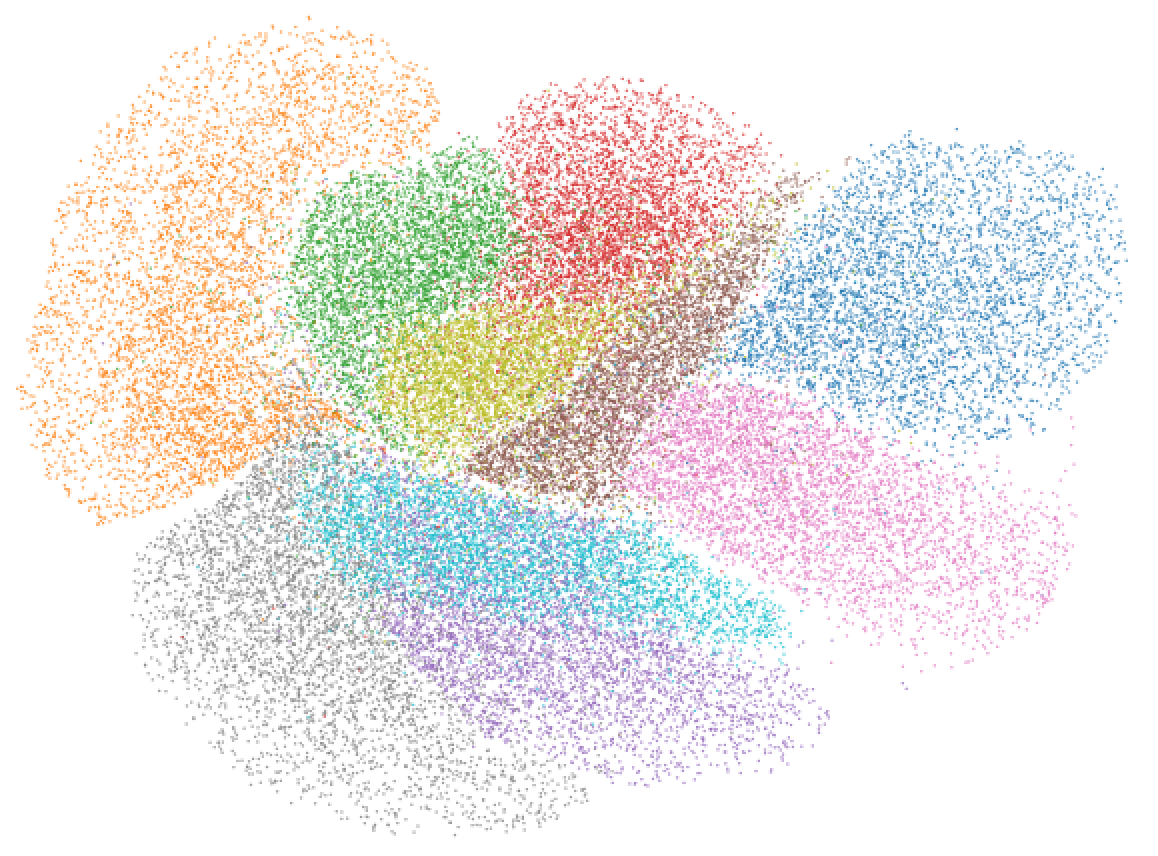}
\end{minipage} &
\begin{minipage}{0.3\textwidth}
    \centering
    \includegraphics[width=\textwidth]{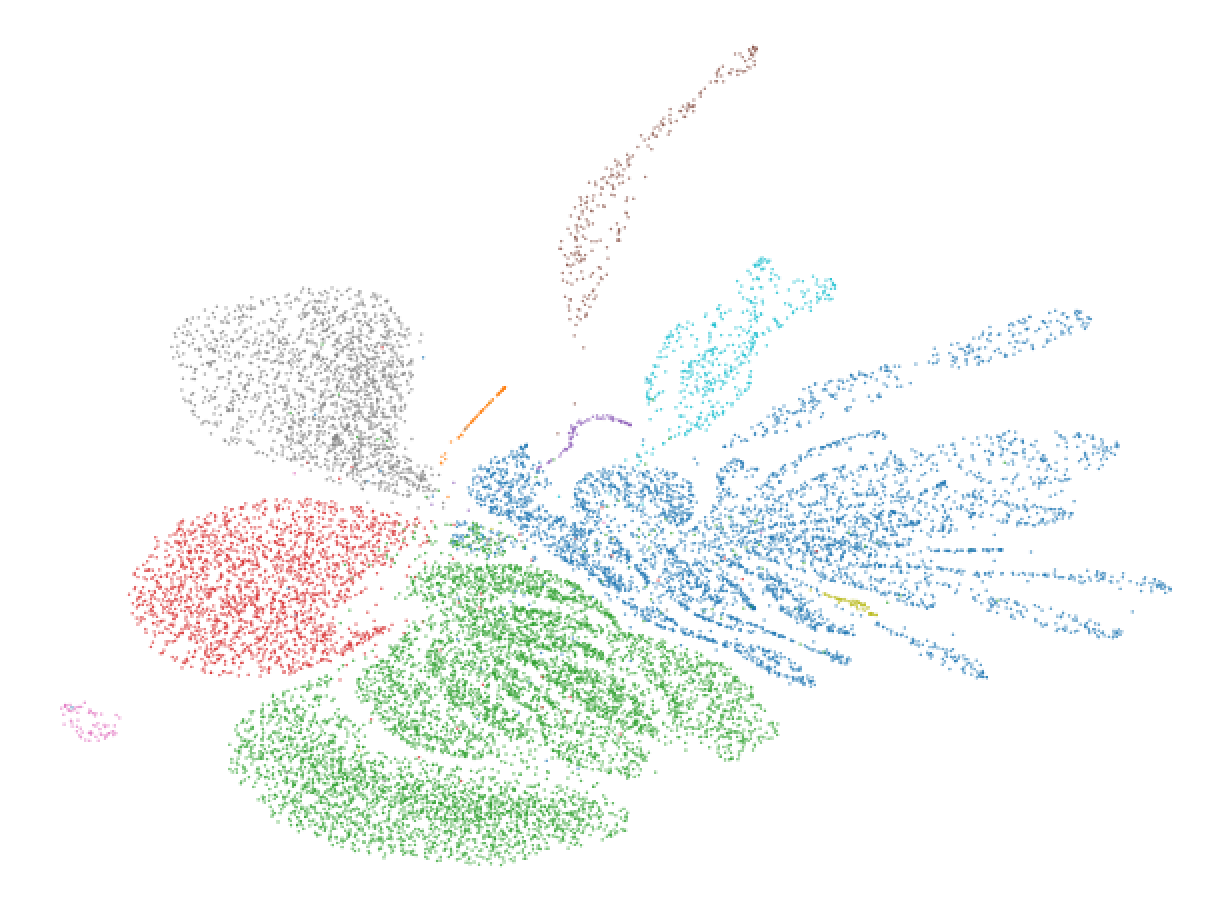}
\end{minipage} &
\begin{minipage}{0.3\textwidth}
    \centering
    \includegraphics[width=\textwidth]{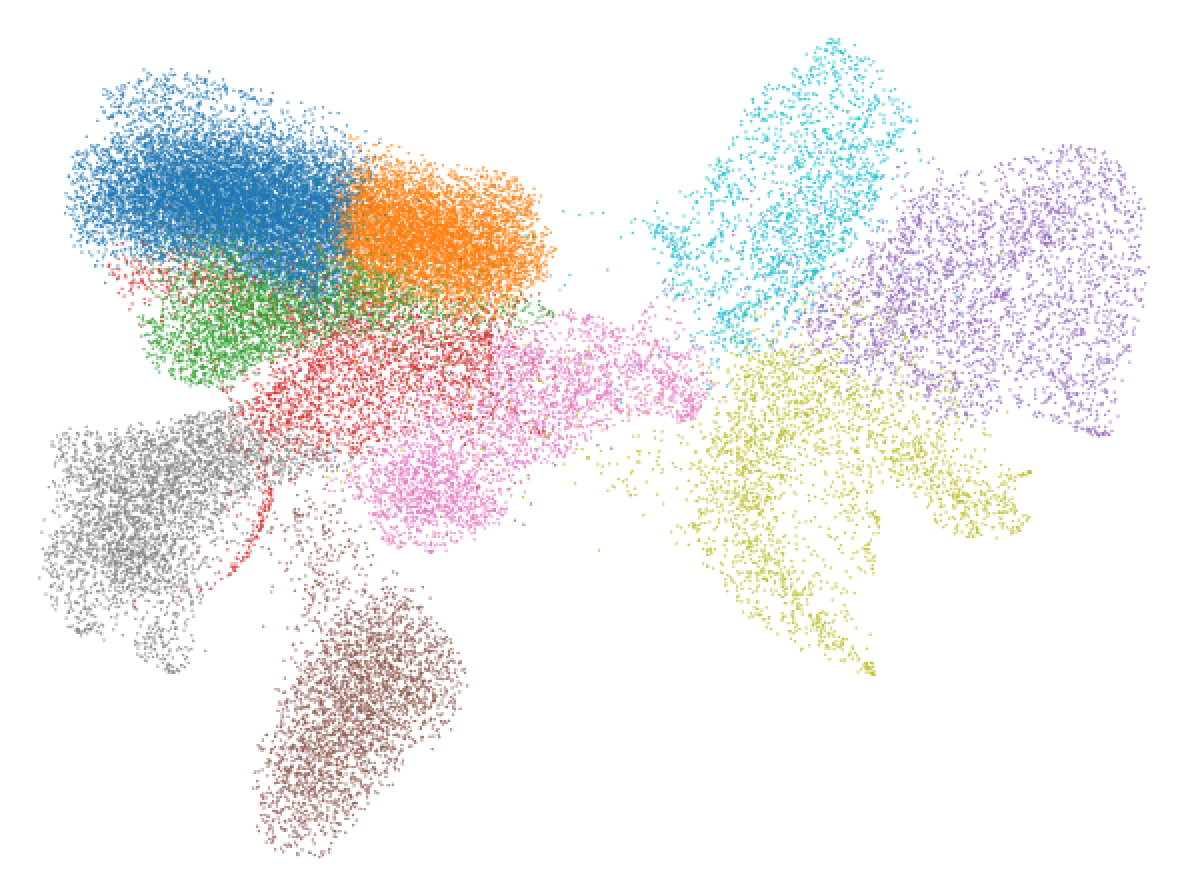}
\end{minipage} \\
\end{tabular}
\caption{Comparison between embeddings obtained using the CNE objective (\textbf{top}) 
and our inference (\textbf{bottom}). 
From left to right: MNIST digits, transcriptomic data from 
\protect\cite{macosko}, and larger-scale transcriptomic data from \protect\cite{zheng}. \cref{app:gplvm} shows that PCA and GPLVMs with a similar kernel do not produce similar embeddings, presumably due to the Laplacian encoding different statistics of the data w.r.t. the empirical covariance.}
\label{fig:all-plots}
\end{figure*}

\cref{fig:all-plots} shows a comparison between embeddings found with our interpretations compared with those presented in \cite{nc_sne} within different datasets (found within the openTSNE repositories \citep{opentsne}). We resample each dataset such that it has exactly 10 groups, with up to 40k points in total. The optimisation was done using a GPU capable of inverting matrices of 40k rows/columns, using pytorch's Adam optimiser, with an initial learning rate set to 1.0, with each experiment being run for 200 epochs, with linear rate decay. We see that the embeddings recovered using our methods are qualitatively quite similar to that of CNE and not PCA/GPLVM.

Note that although we start with a CNE objective that approximates UMAP, our embeddings qualitatively resemble that of neg-t-SNE, and this is the benchmark we compare against. We believe that this is because approximations made in this paper implicitly lower the $\Bar{Z}$ hyperparameter of \cite{nc_sne}. 
A neg-t-SNE based derivation is provided in \cref{app:tsne}. We leave the exploration of tighter bounds and approximations for future work.


The Wishart model of \cref{eqn:wish-interp} is quite elegant, as it implies that the data covariance is modelled by a covariance function, connecting Gaussian process latent variable models \cite{gplvm} (that use kernels not unlike the Student-t/Cauchy kernel to model the covariance) to ProbDR models that describe the graph Laplacian (which describes a precision matrix). Moreover, it is interesting to note that the implied covariance of our model (the inverse of the Wishart parameter) is non-stationary and can be justified by the fact that the adjacency probabilities go to zero as a function of distance. The semantic, probabilistic, and modelling implications behind other assumptions (such as the high noise-scale parameter within the Wishart parameter) made within our simplified ProbDR remain an area that can be studied. In addition to these insights, we show non-traditional ways in which covariances can be constructed as part of Gaussian process models and/or dimensionality reduction methods, which revolve around double-centering CPD matrices (which can be based on distance matrices\footnote{For example, take $-\mathbf{D} = \log \dfrac{1}{1 + \| \X_i - \X_j \|^2}$, where $\mathbf{D}$ represents a distance matrix. The inner term is a kernel, hence is PSD. The $\log$ function (like the square root, in this context) preserves CPSD-ness \citep{psd-mats}. Therefore, $-\H \mathbf{D} \H$ is PSD. It's also interesting to observe, due to the results of \cite{sch-ppt, sch}, that there exists an isometric Euclidean embedding for such a distance matrix (and vice versa).} or element-wise exponentiated kernel matrices\footnote{with fractional powers.}).

\section{Conclusion}

In conclusion, this study presents a novel theoretical framework that reinterprets UMAP and t-SNE-like algorithms as maximum a posteriori (MAP) inference algorithms within a model for a graph Laplacian described by a Wishart distribution. This result bridges the gap between popular dimensionality reduction methods with Gaussian process latent variable models, enhancing our understanding of their mechanisms. The insights gained show the importance of specific modelling assumptions in optimizing these algorithms' effectiveness and interpretability, setting a foundation for further research on model specifications and their practical implications. Next steps include studying whether eigenfunctions of RBF-like kernels (which roughly behave as $\phi_k(x) \sim \gamma \cos \left( 2x\kappa - k\pi/2 \right)$) can aid the characterization of solutions of ProbDR models. Moreover, the unification of DR algorithms presented here may aid in scaling of white-box transformers, introduced by \cite{whitebox-transformer}.

\section*{Acknowledgements}

AR would like to thank Diana Robinson and Francisco Vargas for helpful discussions, and the Accelerate Programme for Scientific Discovery for a studentship funding this work.

%% file: appendix.tex
\section{Approximate distribution of a distance matrix}

\begin{theorem}[Distribution of normal distances]
Assume that $\Y$ is distributed as,
$$\begin{bmatrix} \Y_i \\ \Y_j \end{bmatrix} \sim \mathcal{MN} \left(\mu, \begin{bmatrix} k_{ii} & k_{ij} \\ k_{ji} & k_{jj} \end{bmatrix}, \I_d \right). $$
Then, the following hold. Firstly, denoting $d_{ij}^2 = \| \Y_i - \Y_j\|^2$, the marginal distribution is given by,
$$ d_{ij}^2 \sim \Gamma\left(k=\dfrac{d}{2}, \theta= 2(k_{ii} + k_{jj} - 2k_{ij})\right).$$

As a consequence, $ \mathbb E(d_{ij}^2) = d * \kt_{ij} \text{ and } \mathbb V(d_{ij}^2) = 2d * \kt_{ij}^2$,
where $\kt_{ij} = k_{ii} + k_{jj} - 2k_{ij}$.

Additionally,
$$ \mathbb{C}(d_{ij}^2, d_{mn}^2) = 2d * (k_{im} + k_{jn} - k_{in} - k_{jm})^2. $$
This is a useful fact as the upper triangle of the distance matrix is approximately normal due to the central limit theorem with increasing $d$.
\label{thm:norm_dist}
\end{theorem}

The first part of the theorem is given in ProbDR (Ravuri et al. 2023), reproduced below.
\begin{align*}
    \forall k: d'_{ij} \equiv y_i^k - y_j^k &\sim \mathcal{N} (0, k_{ii} + k_{jj} - 2k_{ij}) \overset{d}{=} \sqrt{k_{ii} + k_{jj} - 2k_{ij}}Z \\
    \Rightarrow d_{ij}^2 \equiv \| \Y_i - \Y_j\|^2 = \sum_k^d (y^k_i - y^k_j)^2 &\overset{d}{=} (k_{ii} + k_{jj} - 2k_{ij}) \sum_k^d Z_k^2 \\
    &\overset{d}{=} (k_{ii} + k_{jj} - 2k_{ij}) \chi^2_d \\
    &\overset{d}{=} \Gamma(k=d/2, \theta=2(k_{ii} + k_{jj} - 2k_{ij})).
\end{align*}

The covariance between $d_{ij}^2$ and $d_{mn}^2$ can be computed as follows. Let,
\begin{align*}
   d'_{ij} = \Y_{id} - \Y_{jd} \;\; \text{and} \;\;
   d'_{mn} = \Y_{md} - \Y_{nd}.
\end{align*}
We can then derive some important moments as follows,
$$ \mathbb{E} (d'_{ij}) = 0, \mathbb{V}(d'_{ij}) = \mathbb{E}(d^{'2}_{ij}) = k_{ii} + k_{jj} - 2k_{ij}, $$

\begin{align*}
\mathbb{C}(d'_{ij}, d'_{mn}) &= \mathbb{C}(Y_{id} - Y_{jd}, Y_{md} - Y_{nd}) \\
&= \mathbb{C}(Y_{id}, Y_{md}) - \mathbb{C}(Y_{id}, Y_{nd}) - \mathbb{C}(Y_{jd}, Y_{md}) + \mathbb{C}(Y_{jd}, Y_{nd}) \\
&= k_{im} + k_{jn} - k_{in} - k_{jm},
\end{align*}

Then,
\begin{align*}
   \mathbb{C}(d_{ij}^2, d_{mn}^2) &= \mathbb{C}\left( \sum_{d_1} (\Y_{id_1} - \Y_{jd_1})^2, \sum_{d_2} (\Y_{md_2} - \Y_{nd_2})^2 \right) \\
   &= \sum_{d_1} \sum_{d_2} \mathbb{C}((\Y_{id_1} - \Y_{jd_1})^2, (\Y_{md_2} - \Y_{nd_2})^2) && \text{linearity} \\
   &= \sum_d \mathbb{C}(d^{'2}_{ij}, d^{'2}_{mn}) && \text{independence} \\
   &= d * \mathbb{C}(d^{'2}_{ij}, d^{'2}_{mn}) \\
   &= d * \left[ \mathbb{E}[d^{'2}_{ij} d^{'2}_{mn}] - \mathbb{E}[d^{'2}_{ij}] \mathbb{E}[d^{'2}_{mn}] \right] \\
   &= d * \left[ \mathbb{E}(d^{'2}_{ij}) \mathbb{E}(d^{'2}_{mn}) + 2\mathbb{E}^2(d'_{ij} d'_{mn}) - \mathbb{E}[d^{'2}_{ij}] \mathbb{E}[d^{'2}_{mn}] \right] && \text{Isserlis' theorem} \\
   &= 2d * \mathbb{E}^2(d'_{ij} d'_{mn}) \\
   &= 2d * (k_{im} + k_{jn} - k_{in} - k_{jm})^2.
\end{align*}

\section{Derivation of \cref{eqn:cne-bound}}
\label{app:derivation-cne-bound}

Eqn. 8 of \cite{nc_sne} reads,
$$\mathcal{L}^{\text{NEG}}(\theta) = -\mathbb{E}_{x\sim p} \log \left( \dfrac{q_\theta(x)}{q_\theta(x) + 1} \right) - m \mathbb E_{x \sim \xi} \log \left( 1 - \dfrac{q_\theta(x)}{q_\theta(x) + 1} \right).$$

We use the notation $n_{\text{neg}} = m$, and we consider the objective (resulting in \cref{eqn:cne-bound}) in terms of the negative loss $\mathcal E = -\mathcal L$. Next, Lemma 3 of \cite{nc_sne} specifies the choice of $q_\theta$ corresponding to UMAP to be $q_\theta(x) = 1/d_{ij}^2(\X)$. The objective simplifies to,
\begin{align*}
\mathcal{E}(\X) &= \dfrac{1}{\sum_{i>j} \A_{ij}} \sum_{i>j} \A_{ij} \log \left( \dfrac{1}{1 + d_{ij}^2(\X)} \right) + \dfrac{n_{\text{neg}}}{\sum_{i>j} (1 - \A_{ij})} \sum_{i>j} (1 - \A_{ij}) \log \left( 1 - \dfrac{1}{1 + d_{ij}^2(\X)} \right) \\
&\propto \sum_{i>j} \A_{ij} \log \left( \dfrac{1}{1 + d_{ij}^2(\X)} \right) + \dfrac{n_{\text{neg}} \sum_{i>j} \A_{ij}}{\sum_{i>j} (1 - \A_{ij})} \sum_{i>j} (1 - \A_{ij}) \log \left( 1 - \dfrac{1}{1 + d_{ij}^2(\X)} \right)
\end{align*}

The multiplicative constant is approximated as,
\begin{align*}
\Tilde{\epsilon} &\equiv \dfrac{n_{\text{neg}}\sum_{i>j}\A_{ij}}{\sum_{i>j}1 - \A_{ij}} \approx \dfrac{n*n_{\text{neg}}* n_{\text{neigh}}/1.5}{(n^2 - n)/2} \approx \dfrac{4n_{\text{neg}} n_{\text{neigh}}}{3n}.
\end{align*}
Therefore, the objective becomes,
\begin{align*}
\mathcal{E}(\X) &\approx \sum_{i>j} \A_{ij} \log \left( \dfrac{1}{1 + d_{ij}^2(\X)} \right) + \dfrac{4n_{\text{neg}} n_{\text{neigh}}}{3n} \sum_{i>j} (1 - \A_{ij}) \log \left( 1 - \dfrac{1}{1 + d_{ij}^2(\X)} \right) \\
&\propto \sum_{ij} \A_{ij} \log \left( \dfrac{1}{1 + d_{ij}^2(\X)} \right) + \dfrac{4n_{\text{neg}} n_{\text{neigh}}}{3n} \sum_{ij} (1 - \A_{ij}) \log \left( 1 - \dfrac{1}{1 + d_{ij}^2(\X)} \right),
\end{align*}
which is \cref{eqn:cne-bound}.

\section{Comparison with GPLVMs}
\label{app:gplvm}

\begin{figure*}[htp]
\centering
\begin{tabular}{ccc}
\begin{minipage}{0.3\textwidth}
    \centering
    \includegraphics[width=\textwidth]{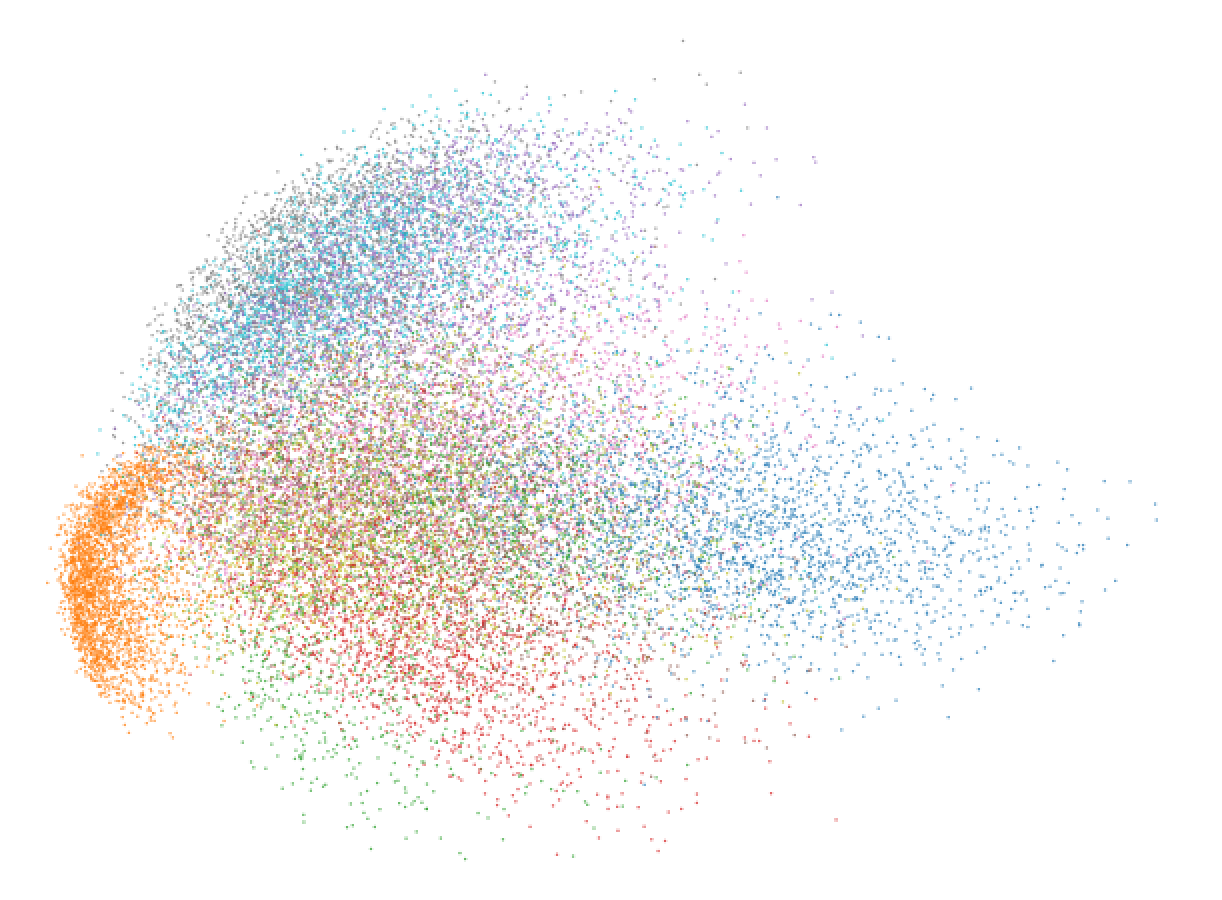}
\end{minipage} &
\begin{minipage}{0.3\textwidth}
    \centering
    \includegraphics[width=\textwidth]{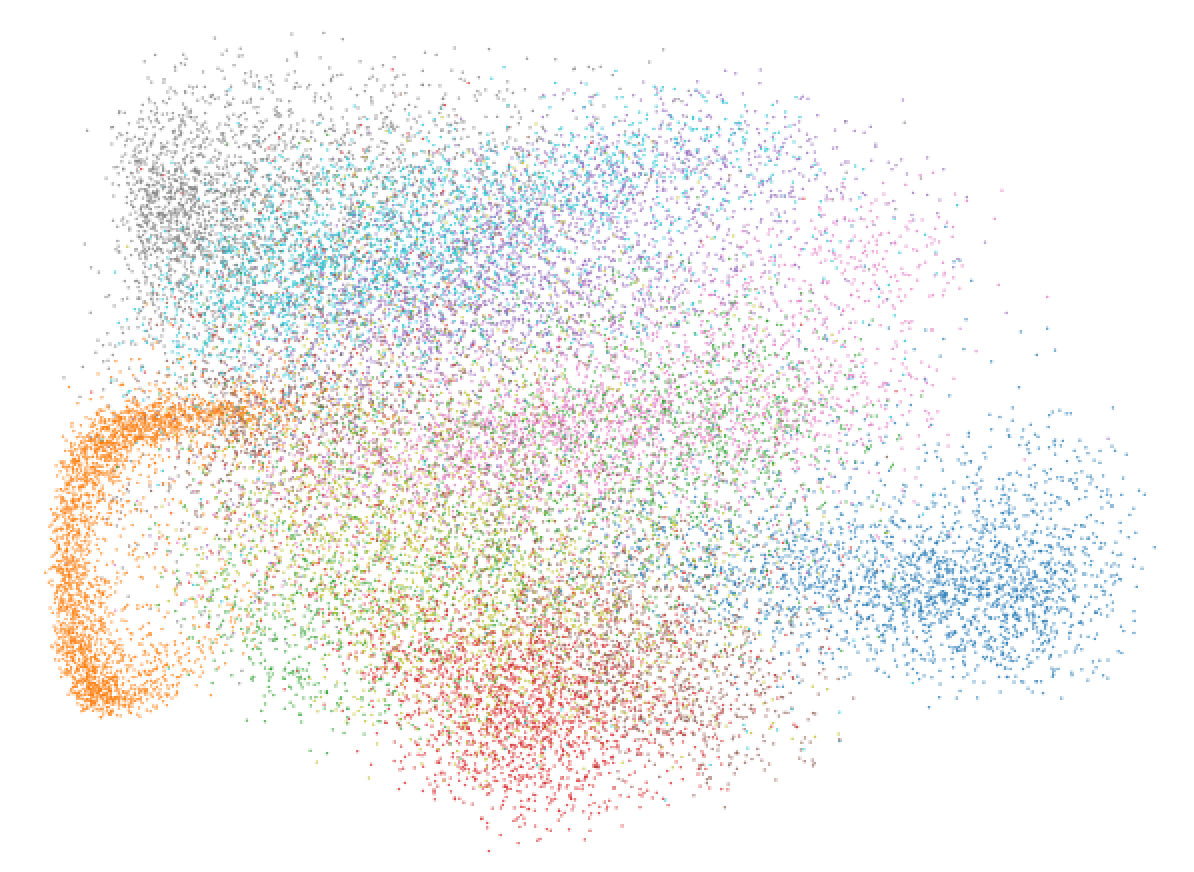}
\end{minipage} &
\begin{minipage}{0.3\textwidth}
    \centering
    \includegraphics[width=\textwidth]{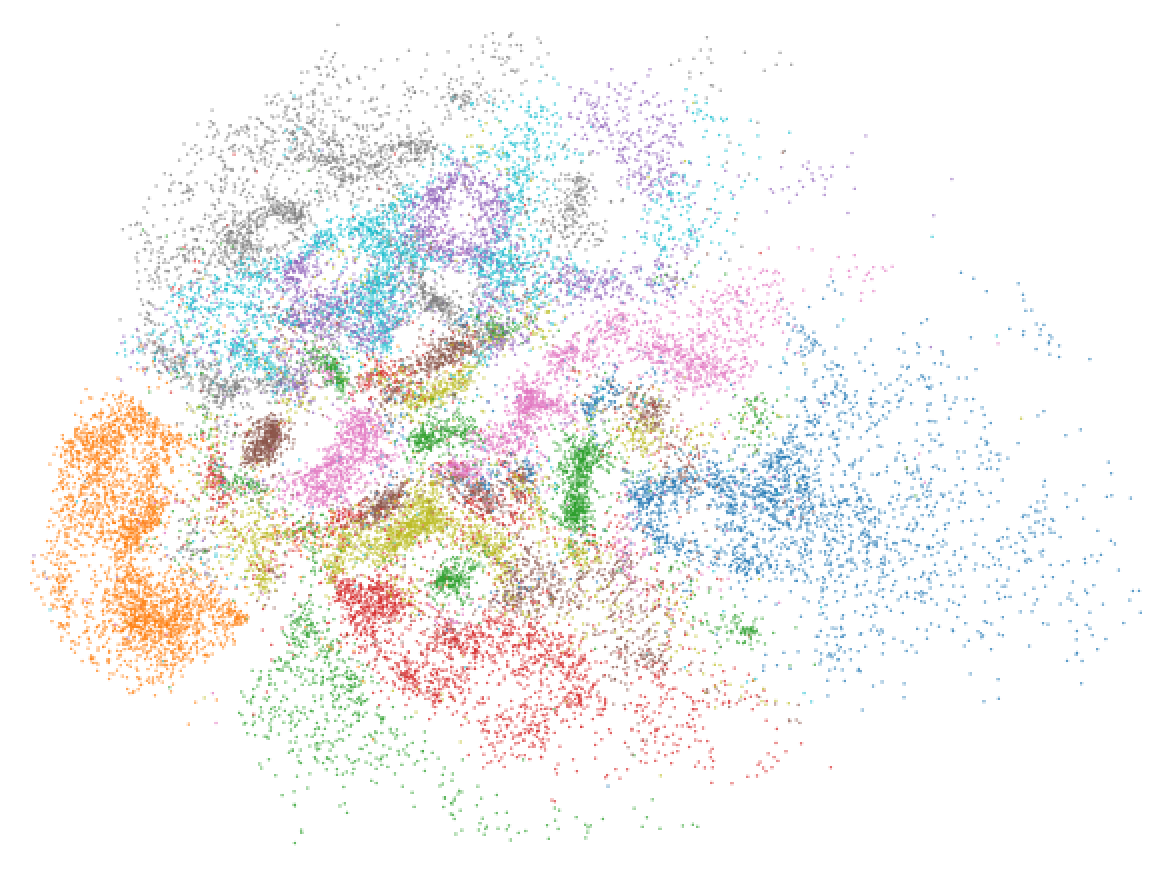}
\end{minipage}
\end{tabular}
\caption{MNIST digits embedded using PCA (\textbf{left}), GPLVM using a \texttt{linear + constant + t + noise} kernels, with the inits scaled towards zero (\textbf{center}), and the same GPLVM with unscaled PCA inits (\textbf{right}). In each case, the GPLVM hyperparameters were first ``pre-trained'' using the PCA-initialized embeddings for 10 epochs, and the embeddings were trained for a further 40. In every case, note the visual dissimilarity w.r.t. our versions of UMAP/t-SNE.}
\label{fig:app-gplvm}
\end{figure*}

A comparison with GPLVMs using similar kernels to ours is shown in \cref{fig:app-gplvm}.

\section{Using neg-t-SNE as a starting point}
\label{app:tsne}

Through an ablation of the implementation of \cite{nc_sne}, we find that the objective neg-t-SNE (an approximation to t-SNE), maximises,
\begin{align*}
    \mathcal{E}(\X) \propto &\sum_{ij} \A_{ij} \log\left(\frac{1}{1 + \Tilde{s}(d_{ij}(\X)^2 + 1)}\right) + \dfrac{4n_{\text{neg}} n_{\text{neigh}}}{3n}\sum_{ij} (1 - \A_{ij}) \log\left(1 - \frac{1}{1 + \Tilde{s}(d_{ij}(\X)^2 + 1)}\right),
\end{align*}
after first pretraining the embedding with the UMAP-approximating objective of \cref{eqn:cne-bound} for a third of the overall number of iterations. The ``spec param'' in the code is $\Tilde{s} = 100n_{\text{neg}}/n$.

We follow the same steps as our initial derivation, with $d_{ij}^2 \mapsto \Tilde{s}(d_{ij}^2 + 1)$,
$$ \Tilde{\epsilon} \log \left( 1 - \dfrac{1}{1 + \Tilde{s}(d_{ij}^2 + 1)} \right) \approx \log \left(1 - \Tilde{\epsilon} \log \left(1 + \dfrac{1}{\Tilde{s}(d_{ij}^2 + 1)} \right) \right) \approx \log \left(1 - \Tilde{\epsilon} \log \left(1 + \dfrac{1}{\Tilde{s}d_{ij}^2} \right) \right). $$

Using an ansatz $\log(1+1/x) \approx c/(1+x)$, we approximate this term as,
$$ \log \left(1 - \Tilde{\epsilon} \log \left(1 + \dfrac{1}{\Tilde{s}d_{ij}^2} \right) \right) \approx \log \left( 1 - \dfrac{\Tilde{\epsilon}\log(1 + \Tilde{s}^{-1})}{1 + \Tilde{s}d_{ij}^2} \right). $$

Even though the approximation is quite coarse, we note that empirically, the objectives result in similar embeddings. Therefore, following our initial derivation, the neg-t-SNE objective is approximated by,
\begin{align*}
    \mathcal{E} \approx &\sum_{ij} \A_{ij} \log\left(\frac{\Tilde{\epsilon}\log(1 + \Tilde{s}^{-1})}{1 + \Tilde{s}d_{ij}(\X)^2}\right) + \sum_{ij} \frac{\Tilde{\epsilon}\log(1 + \Tilde{s}^{-1})}{1 + \Tilde{s}d_{ij}(\X)^2}.
\end{align*}

Following the derivation now is straightforward, we find that the neg-t-SNE-based interpretation is,
\begin{align}
    \L/\log1p(\Tilde{s}^{-1}) | \X \sim \mathcal{W} \left((0.5\Tilde{\epsilon}^{-1}\I + 0.5 \H \P^t \H + \Tilde{s} \X \X^T)^{-1}, n \right),
\end{align}
where $\P^t_{ij} = 1/(1 + \Tilde{s}d_{ij}^2)$. Note that, apart from a minor rescaling of the graph Laplacian, the model does not significantly change (the latents are simply rescaled by $\sqrt{\Tilde{s}}$). The resulting embedding from such an embedding is shown below, providing more evidence for our hypothesis that the hyperparameter $\Bar{Z}$ of \cite{nc_sne} may be implicitly lowered by our approximations.

\begin{figure}
    \centering
    \includegraphics[width=0.5\linewidth]{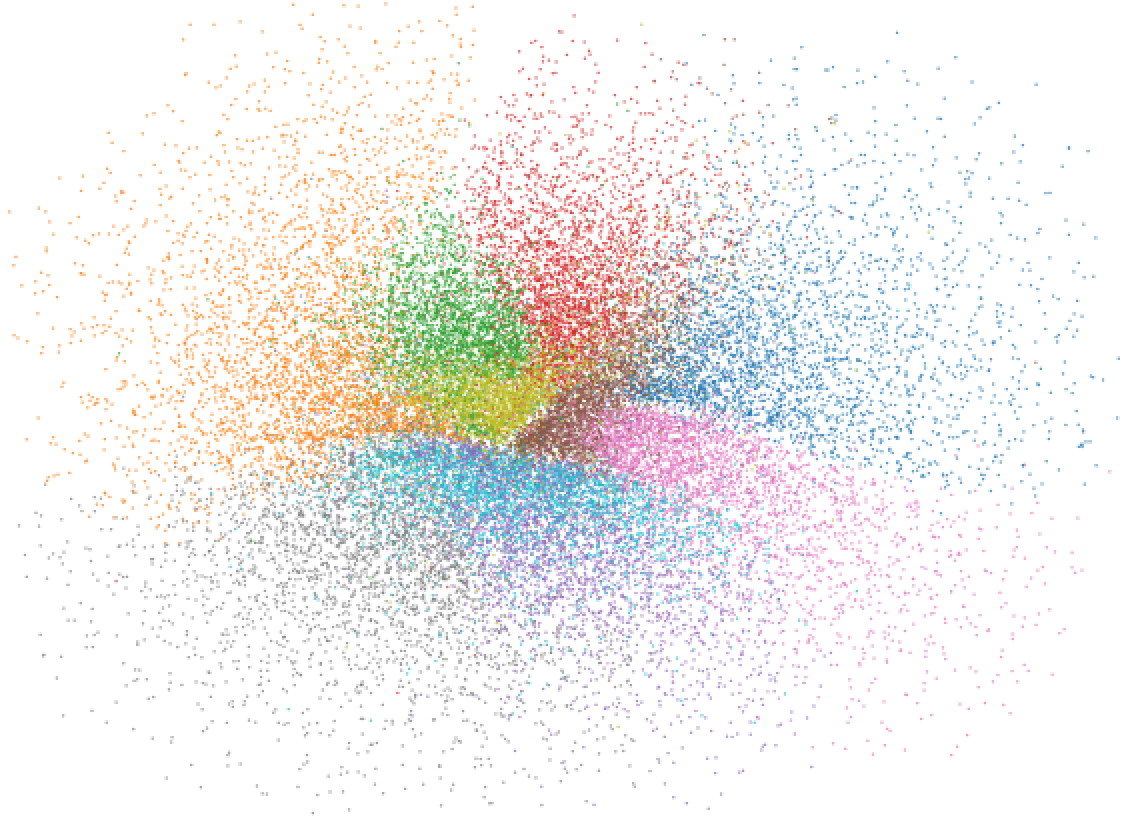}
    \caption{MNIST digits embedded using our probabilistic interpretation derived using neg-t-SNE.}
    \label{fig:tsne-interp}
\end{figure}